\documentclass[a4paper,fleqn]{cas-dc}
\usepackage{graphicx}%
\usepackage{caption} %
\usepackage{multirow}%
\usepackage{amsmath,amssymb,amsfonts}%
\usepackage{amsthm}%
\usepackage{mathrsfs}%
\usepackage[title]{appendix}%
\usepackage[table,xcdraw]{xcolor}%
\usepackage{textcomp}%
\usepackage{manyfoot}%
\usepackage{booktabs}%
\usepackage{algorithm}%
\usepackage{algorithmicx}%
\usepackage{algpseudocode}%
\usepackage{listings}
\usepackage{setspace}
\usepackage{adjustbox}
\usepackage{float}
\usepackage{doi}
\usepackage{hyperref}
\usepackage{tabularx}
\usepackage{array}
\usepackage{float}

\makeatletter
\@ifundefined{orcidlogo}{}{%
  
}
\makeatother
\usepackage{orcidlink}

\floatstyle{plaintop}
\restylefloat{table}
\usepackage[numbers,sort&compress]{natbib}

\setcitestyle{numbers,square,comma,sort&compress}

\bibpunct{[}{]}{,}{n}{}{,} 

\theoremstyle{thmstyleone}%

\theoremstyle{thmstyletwo}%

\theoremstyle{thmstylethree}%
\def\tsc#1{\csdef{#1}{\textsc{\lowercase{#1}}\xspace}}
\tsc{HM}
\tsc{GR}
\tsc{MS}
\tsc{MH}

\begin{document}
\let\WriteBookmarks\relax
\def\floatpagepagefraction{1}
\def\textpagefraction{.001}
\shorttitle{Translation as a Computationally Efficient Bridge}
\shortauthors{H. Muizelaar et~al.}

\title[mode = title]{Translation as a Computationally Efficient Bridge: Feasibility of English BERT for Low-Resource Languages}                    
   
\author[1]{Hielke Muizelaar}[type=editor,
                        auid=000,bioid=1,
                        orcid=0009-0002-3593-7824]
\cormark[1]
\ead{h.j.muizelaar@lumc.nl}
\credit{Conceptualization, Methodology, Project administration and supervision, Resources, Writing - Original draft preparation, Writing - Reviewing and editing}

\affiliation[1]{organization={Department of Public Health and Primary Care, Leiden University Medical Center},
                addressline={Albinusdreef 2}, 
                city={Leiden},
                postcode={2333ZA}, 
                country={The Netherlands}}

\author[2]{Giulia Rivetti}[type=editor,
                        auid=000,bioid=1]
\ead{giulia.rivets@gmail.com}
\credit{Data curation, Methodology, Software, Experiments, Result Validation, Visualizations, Writing - Reviewing and editing}

\affiliation[2]{organization={LIACS, Leiden University},
                addressline={P.O. Box 9512}, 
                city={Leiden},
                postcode={2300RA}, 
                country={The Netherlands}}

\author[1,2]{Marco Spruit}[type=editor,
                        auid=000,bioid=1,
                        orcid=0000-0002-9237-221X]
\ead{m.r.spruit@lumc.nl}
\credit{Conceptualization, Project administration and supervision, Resources, Writing - Reviewing and editing}

\author[1]{Marcel Haas}[type=editor,
                        auid=000,bioid=1,
                        orcid=0000-0003-2581-8370]
\ead{m.r.haas@lumc.nl}
\credit{Conceptualization, Methodology, Project administration and supervision, Resources, Writing - Reviewing and editing}
\begin{abstract}
BERT models have revolutionised Natural Language Processing (NLP) through their ability to process unstructured text across diverse domains. However, developing high-quality BERT models for non-English languages remains challenging due to limited annotated data and high computational demands. Translating non-English data into English and fine-tuning existing English BERT models offers a resource-efficient alternative, yet few studies have structurally compared translation-based fine-tuning with native-language BERT performance across tasks and languages. This study provides such a comparison, evaluating the feasibility of translation-based fine-tuning across six NLP tasks: Sentiment Analysis, Hate Speech Detection, Question Answering, Named Entity Recognition, Part-of-Speech Tagging, and Natural Language Inference, using datasets translated from Bulgarian, Chinese, Dutch, Italian, and Russian. Across all settings, the translation-based approach was comparable or superior in 53.3 percent of cases. Gains were most frequent in Question Answering, Part-of-Speech Tagging, and Natural Language Inference, while performance declines were common in Named Entity Recognition and Hate Speech Detection. The results show that translation-based fine-tuning is most effective for tasks relying on syntactic or structural patterns and for languages typologically close to English, such as Dutch, but less effective for token-level or culturally nuanced tasks, particularly in Chinese. Overall, this study demonstrates that translation-based fine-tuning offers a scalable, resource-efficient, and empirically validated path for extending NLP to low-resource languages while advancing linguistic inclusivity and sustainability in artificial intelligence.
\end{abstract}


\maketitle

\section{Introduction}





In recent years, advances in Natural Language Processing (NLP) have transformed the way machines process, understand, and generate human language. This progress has been driven primarily by transformer-based architectures, beginning with models such as BERT (Bidirectional Encoder Representations from Transformers) \citep{bert}, which set new standards across a wide range of NLP tasks. Subsequent models, including RoBERTa \citep{roberta}, XLM-R \citep{xlmr}, and more recent large language models (LLMs) \citep{cit_llmsurvey, cit_gptsurvey, cit_gpt4}, have further advanced the state-of-the-art. However, these developments remain unevenly distributed across languages. English and a limited number of other high-resource languages benefit from abundant annotated data and large corpora, whereas many languages remain under-represented in NLP research and tools. 

This imbalance creates a practical challenge for multilingual NLP: researchers and practitioners must decide whether to develop or fine-tune language-specific models, rely on multilingual encoders, use generative LLMs through prompting, or translate data into a high-resource language, and apply mature English-language models. Each strategy involves trade-offs. Language-specific models may offer strong performance, but require suitable corpora, annotated data, and computational resources \citep{joshi, ct144, cit_energy}. Multilingual encoders such as mBERT, XLM-R, and mDeBERTa provide broader coverage but may underperform for specific languages, domains, or tasks \citep{ct144, ct47, litschko}. Generative LLMs increasingly support multilingual zero-shot and few-shot inference, including cross-lingual transfer across diverse tasks and languages \citep{lin2022fewshot,asai2024buffet}. However, their use also raises practical and methodological concerns related to cost, access restrictions, reproducibility, prompt sensitivity, and structured-output reliability, especially for token-level tasks such as named entity recognition \citep{bommasani2021foundation,ji2023vicunaner,polo2024prompteval,sivarajkumar2024prompting,shorten2024structuredrag,lu2024tokenlevelner}. Translation-based fine-tuning, therefore, remains a relevant and resource-efficient alternative, but its reliability across languages and NLP tasks is still insufficiently characterised.  

A central obstacle in low-resource NLP is the lack of annotated datasets for supervised learning. Joshi et al. note that many languages have little or no digital presence, making the creation of robust NLP systems extremely challenging \cite{joshi}. The issue is not limited to rarely spoken languages: even populous languages such as Spanish and Russian face gaps in specific domains, including biomedical NLP and social media sentiment analysis \citep{ct17, ct14, ct16, ct65}. Morphologically rich and structurally complex languages, including Turkish \citep{ct109}, Thai \citep{ct67, ct56}, and Amharic \citep{ct149}, further complicate cross-lingual transfer and pre-training, as their linguistic characteristics are poorly captured by existing models. These challenges also vary between language families, since the typological distance from English often reduces transfer effectiveness.

Multilingual models such as mBERT have been proposed to alleviate these limitations, but their performance is often suboptimal for truly under-represented languages. Wu and Dredze showed that the performance of mBERT decreases drastically for languages with minimal training data \citep{ct144}. Even monolingual models built for these languages often fail to reach competitive accuracy without extensive pre-training or domain-specific corpora \citep{ct72}. This situation is further complicated in tasks such as intent detection and slot filling, where data labelling remains expensive and inconsistent \citep{ct112}. More recent multilingual encoders and instruction-tuned LLMs have improved cross-lingual capabilities, but do not remove the need to understand when simpler encoder-based translation pipelines remain competitive, particularly in settings where local execution, reproducibility, and limited computational resources are crucial constraints.

Training language-specific BERT models from scratch is one solution, but this is highly resource intensive and requires significant computational power, specialised hardware, and very large raw text corpora for pre-training \citep{cit_costnlp, cit_energy, cit_extremebert}. To address these challenges, an increasing number of studies have explored the potential of using machine translation. By translating text from low-resource languages into English, researchers can take advantage of well-developed English-language models such as BERT. This method has shown encouraging results in several NLP tasks, suggesting that it may serve as a practical alternative to developing language-specific models. For sentiment analysis, Balahur et al. showed that machine-translated corpora achieved accuracy within 8\% of native-language models, showing the viability of this approach \cite{ct10}. Refaee and Rieser further showed that translated Arabic tweets, when analysed using English sentiment tools, could outperform native baselines, illustrating that English resources can be extended to support under-represented languages \cite{ct106}. Similarly, Muizelaar et al. demonstrated that translating Dutch clinical texts into English and fine-tuning pre-trained English clinical BERT models produced comparable or superior performance to Dutch clinical BERT on the task of lifestyle-related text classification \cite{muiz}.

At the same time, translation can introduce errors that affect downstream performance. Generative and lexical alignment tasks, for instance, are particularly sensitive to translation errors. For example, Meng et al. observed that bilingual sentiment lexicons derived from machine translation often suffered from limited coverage and ambiguity \cite{ct77}. In biomedical NLP, Dorendahl et al. found that applying English tools such as MetaMap to German texts translated by machine translation produced suboptimal results \cite{ct30}. Even translation-based dataset augmentation can fail: Demirtas et al. showed that translation-induced inaccuracies prevented performance improvements in sentiment classification \cite{ct26}. Recent work on LLM-based machine translation further shows that translation quality can strongly depend on inference settings, prompting strategies, and language direction \citep{ctpeng, ctzhang, ctfarinhas}. These findings underscore that translation should not be treated as a neutral preprocessing step: translation artefacts, hallucinations, off-target output, omissions, and alignment loss can all influence downstream NLP performance. 

Although generative LLMs provide a powerful alternative for multilingual NLP, they address a different practical and experimental setting from the one evaluated here. This study focuses on a common use case in multilingual NLP: applying supervised encoder-based models to large bodies of text in non-English and lower-resource languages. In this setting, training data, labels, tokenisation, and evaluation metrics can be kept constant across languages and tasks, enabling a controlled comparison of translation-based fine-tuning with language-specific BERT models. In contrast, LLM prompting introduces additional sources of variation, including prompt design, decoding parameters, model access, model versioning, output-format control, and per-token or API-based inference costs that can become substantial when processing large volumes of text. These issues are especially relevant for structured tasks, where predictions must align with token- or span-level annotations rather than free-form text. Therefore, we treat LLM-based prompting as a complementary strategy rather than as the direct comparator for this study. Instead, we ask under what conditions an open-source translate-train pipeline can provide a viable alternative. By evaluating a transparent, locally executable, and reproducible encoder-based strategy, this study addresses settings where large volumes of text need to be processed or where annotated data, computational resources, and access to commercial or very large models are limited.

Although previous studies have provided valuable information, most have focused on narrow use cases or single-language evaluations, leaving open questions about how translation-based fine-tuning performs across different linguistic families and NLP task types. In particular, sentence-level tasks such as sentiment analysis, hate speech detection, and natural language inference may be less affected by translation than token-level tasks such as named entity recognition and part-of-speech tagging, where token boundaries and label alignment can change after translation. Similarly, languages typologically closer to English may benefit more from translation than structurally distant languages, although dataset quality and task design may also influence performance.  

We investigate whether translating datasets from multiple language families into English and fine-tuning a pre-trained English BERT model can deliver performance comparable to or better than language-specific BERT models for Bulgarian, Chinese, Dutch, Italian, and Russian. These languages were chosen to represent various linguistic families and typological characteristics: Slavic (Bulgarian, Russian), Germanic (Dutch), Romance (Italian), and Sino-Tibetan (Chinese). To capture a broad spectrum of NLP challenges, we evaluated six tasks: sentiment analysis (SA), hate speech detection (HSD), natural language inference (NLI), question answering (QA), named entity recognition (NER), and part-of-speech tagging (POS). Because token-level tasks are particularly sensitive to translation-induced alignment loss, we interpret the POS and NER results separately from sentence-level classification and inference tasks, and explicitly discuss the limitations introduced by automatic re-labelling or alignment procedures.

The contributions of this study are threefold. First, we provide a multi-task, multi-language benchmark of an open-source translate-train pipeline using OPUS-MT and English BERT. Second, we analyse how the success of this pipeline varies across task types, distinguishing sentence-level tasks that are relatively robust to translation from token-level tasks that are more vulnerable to alignment loss. Third, we examine cross-linguistic patterns across five typologically diverse languages, showing how translation-based performance depends on the interaction between linguistic proximity, dataset characteristics, task sensitivity, and translation fidelity. 

\section{Methods}

\subsection{Data}

In this section, we provide an overview of the employed datasets and outline the preprocessing steps that are necessary for the tasks at hand.

\subsubsection{Dataset Selection}

The choice of datasets is crucial to ensure a fair and representative evaluation across tasks and languages. In this study, we selected datasets based on size, structure, and availability of detailed documentation about their creation and intended use. Whenever possible, we relied on established and widely used resources in order to facilitate comparability with prior work and to ensure reproducibility.

For POS tagging and NER, we use datasets from the XTREME (Cross-lingual TRansfer Evaluation of Multilingual Encoders) benchmark \citep{ct47}. XTREME is one of the most widely adopted resources for evaluating multilingual NLP models, covering a diverse set of tasks and languages. Its POS and NER datasets are particularly well suited to our study because they are available for all five languages considered (Bulgarian, Chinese, Dutch, Italian, and Russian) and have been preprocessed and standardised for cross-lingual comparison. Using XTREME ensures that our evaluation is both consistent across languages and directly comparable to previous cross-lingual research.

For the remaining tasks; SA, HSD, QA, and NLI, datasets were selected individually for each language, prioritising availability, prior use in the literature, and suitability for fine-tuning transformer-based models. An overview of all selected datasets is provided in Table~\ref{tab:datasets_overview}, which reports the number of rows per task and language. Additional details, including label distributions of the classification tasks, are provided in Appendix A.

\renewcommand{\arraystretch}{1.15}

\begin{table*}[t] \centering \footnotesize \renewcommand{\arraystretch}{1.15} \setlength{\tabcolsep}{3pt} \begin{tabularx}{\textwidth}{@{}lXXXXX@{}} \toprule \textit{Task} & \textit{Dutch} & \textit{Italian} & \textit{Bulgarian} & \textit{Russian} & \textit{Chinese} \\ \midrule POS & Alpino [D1] (208,747) & PoSTWITA-UD [I1] (129,668) & UD Bulgarian-BTB [B1] (156,149) & Taiga [R1] (197,001) & UD Chinese-GSD [C1] (123,291) \\ NER & wikiann-nl [D2] (40,000) & wikiann-it [I2] (40,000) & wikiann-bg [B2] (40,000) & wikiann-ru [R2] (40,000) & wikiann-zh [C2] (40,000) \\ SA & DBRD [D3] (21,895) & Italian Tweets [I3] (165,815) & Cinexio Movie Reviews [B3] (9,827) & RuReviews [R3] (180,000) & Weibo Senti 100k [C3] (119,988) \\ HSD & Dutch HateCheck [D4] (3,765) & Multilingual Hate Speech [I4] (6,839) & Hate Speech Classification in Bulgarian [B4] (102,750) & Russian South Park [R4] (15,875) & COLD [C4] (37,480) \\ NLI & SICK-NL [D5] (9,840) & LingNLI [I5] (34,878) & XNLI-bg [B5] (400,202) & XNLI-ru [R5] (400,202) & XNLI-zh [C5] (400,202) \\ QA & P-Direkt [D6] (323) & QA-ITA-200k [I6] (202,471) & EXAMS [B6] (3,349) & SberQuAD [R6] (74,300) & CMRC2018 [C6] (14,363) \\ \bottomrule \end{tabularx} \vspace{0.5em} \begin{minipage}{\textwidth} \footnotesize [D1] \citep{cit_posnl}, [I1] \citep{cit_posit}, [B1] \citep{cit_posbg}, [R1] \citep{cit_posru}, [C1] \citep{cit_poszh}. [D2], [I2], [B2], [R2], [C2] all use the WikiAnn datasets \citep{cit_ner}. [D3] \citep{cit_sanl}, [I3] \citep{cit_sait}, [B3] \citep{cit_sabg}, [R3] \citep{cit_saru}, [C3] \citep{cit_sazh}. [D4] \citep{cit_hsdnl}, [I4] \citep{cit_hsdit}, [B4] \citep{cit_hsdbg}, [R4] \citep{cit_hsdru}, [C4] \citep{cit_hsdzh}. [D5] \citep{cit_nlinl}, [I5] \citep{cit_nliit}. [B5], [R5], [C5] all use the XNLI datasets \citep{ct47}. [D6] \citep{cit_qanl}, [I6] \citep{cit_qait}, [B6] \citep{cit_qabg}, [R6] \citep{cit_qaru}, [C6] \citep{cit_qazh}. \end{minipage} \caption{Dataset names and sizes for each NLP task and language.} \label{tab:datasets_overview} \end{table*}
\subsubsection{Data Preprocessing}
\begin{figure*}[t] 
    \centering
    \includegraphics[width=\linewidth]{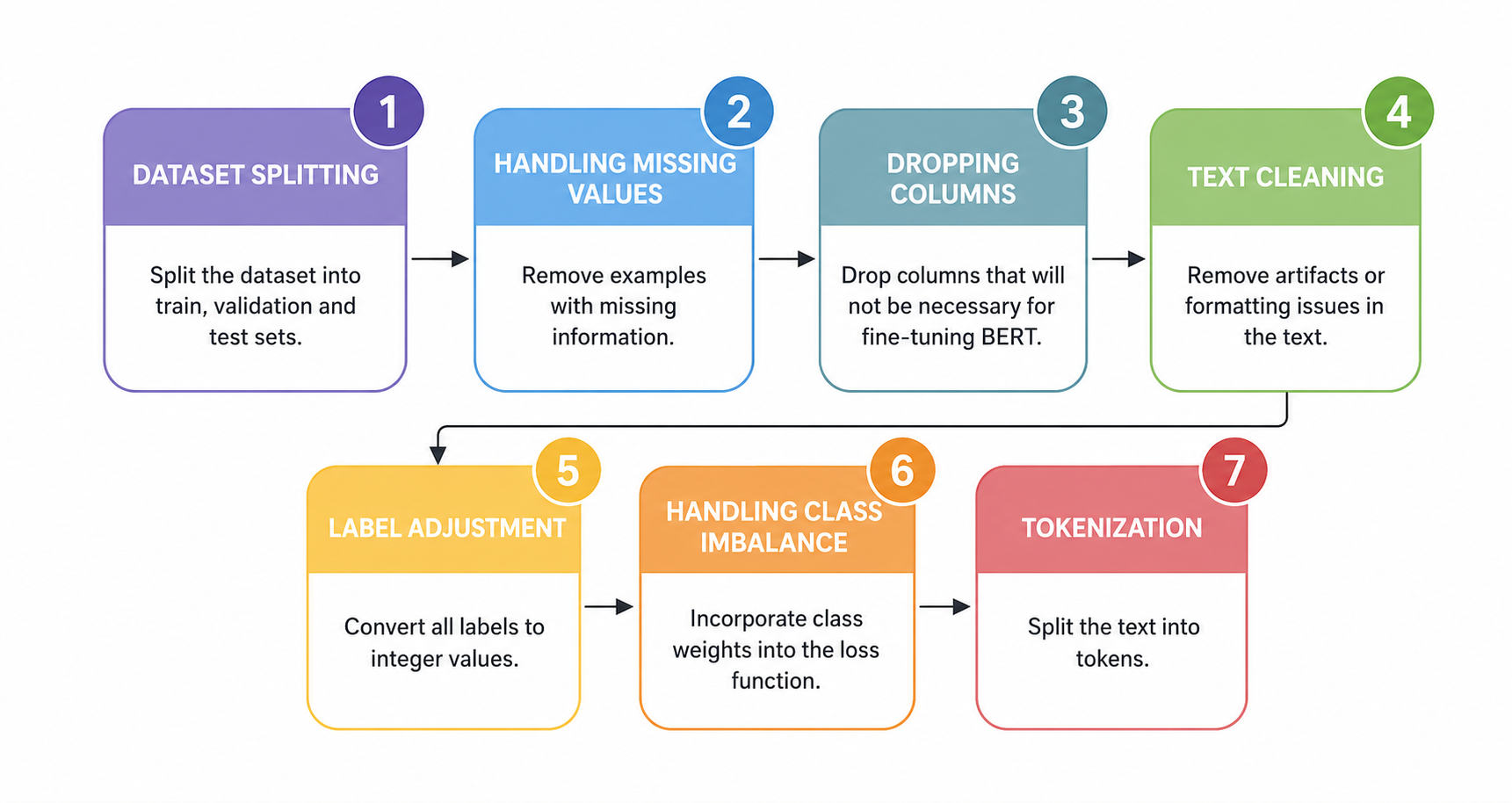} 
    \caption{Overview of the general preprocessing pipeline applied before model fine-tuning. The pipeline includes dataset splitting, removal of incomplete examples, dropping non-essential columns, text cleaning, label conversion, class-imbalance handling, and tokenisation. Task-specific preprocessing steps, such as answer-span realignment for question answering and POS re-labelling after translation, are described separately in the text.}
    \label{fig:pipeline} 
\end{figure*}

The preprocessing pipeline adopted in this work is illustrated in Figure \ref{fig:pipeline} and consists of the following key steps:

\begin{enumerate}  
    \item \textit{Dataset splitting:} We followed official dataset splits. When official splits were not provided, each dataset was randomly divided into training (80\%), validation (10\%), and test (10\%) subsets.  
    \item \textit{Handling missing values:} Examples containing missing or corrupted values, whether originally present or introduced during translation. Corrupted values were identified when Hugging Face models were unable to parse the input during tokenisation or model loading.  
    \item \textit{Dropping unnecessary columns:} Non-essential metadata columns that were not required for BERT fine-tuning were discarded.  
    \item \textit{Text cleaning:} Dataset-specific cleaning procedures were applied, such as removing unwanted characters or artefacts. For NER, we removed spurious punctuation marks introduced during word-by-word translation. For question answering, we reformat all datasets to match the structure of the SQuAD dataset, in order to have a consistent format for questions, contexts, and answers. In the cases of HSD and SA, we apply a more intensive data cleaning process, where we perform lowercasing, removing URLs and special characters, removing extra whitespace, removing non-standard punctuation, removing usernames and handling missing values by removing incomplete samples. 
    \item \textit{Label adjustment:} Labels were reviewed and, where necessary, converted into integer format to ensure compatibility with training. For token-level tasks (NER, POS), labels were aligned with subword tokenization by assigning the entity or POS tag to the first subword and propagating it to the rest.  
    \item \textit{Class imbalance handling:} Where class distributions were highly skewed, balancing techniques were applied to mitigate bias during training. For example, we compute class weights and incorporate these into the loss function during training for imbalanced datasets. These class weights are calculated based on class distribution, where less frequent classes get a higher weight to prevent the models from simply optimising for the most frequent class. This strategy is applied to datasets where the class distribution ratio is 60:40 or more imbalanced in order to improve model generalisation and robustness.   
    \item \textit{Tokenisation:} All text was tokenised with the appropriate BERT tokeniser. For sentence-level classification tasks (SA, HSD, NLI), labels were assigned at the sequence level. For NER and POS tagging, word-level labels were aligned with the tokenised input. For NLI, the tokenizer processed the premise-hypothesis pairs with the \textit{[SEP]} token to capture between-sentence dependencies. For QA, if the context is too long to fit within the model's maximum input length, it is divided into overlapping chunks using a sliding-window mechanism known as stride in order to ensure potential answer spans are not truncated at chunk boundaries. After tokenisation, we track the original example each tokenised chunk corresponds to and align token indices with the character-level answer span to correctly map the start and end positions of the answer within the tokenised input.  
\end{enumerate}  

In addition to this pipeline which is applied across all tasks, certain task-specific challenges require specialised handling. 
For \textit{QA}, an essential processing step in translated texts is to recompute the answer start index within the translated context. Translation often alters the structure of the sentences, the order of words, or the phrasing, invalidating the original indices defined in the source language. To address this, we use Hugging Face's \textit{transformers} library and apply a QA pipeline built on a pretrained DistilBERT model \citep{distilbert}. For each translation question-context pair, the pipeline predicts the most likely answer span, from which both the answer text and its new starting index were extracted. These recomputed values are then used to fine-tune the BERT-Base model. In cases where the model fails to return a valid result, the samples are discarded.
Some QA datasets, such as \textit{EXAMS}, \textit{QA-ITA-200k}, and \textit{CMRC2018}, are missing answer start indices in their original form. For these, we first attempt a direct substring match of the answer within the context. When that fails, we apply fuzzy string matching to locate a similar phrase in the context. If a match is found above a similarity threshold, we compute the corresponding start and end indices. This approach achieves full alignment for \textit{CMRC2018} and good coverage in \textit{EXAMS}, where the answer start index is found for 92\% of the entries. It does, however, prove ineffective for \textit{QA-ITA-200k}, where approximately 78\% of examples lack a detectable match. As a result, we adopt the same strategy used for translated datasets: applying a QA pipeline based on the XLM-RoBERTa model to infer the answer span and starting index. When using the QA pipeline, so for the cases of the translated datasets and the Italian original dataset, the method always predicts a valid answer plan, allowing us to recompute and replace the original annotations in 100\% of cases. In these scenarios, no examples are discarded. However, it is important to note that while the pipeline returns consistent results, the correctness of these predictions is not guaranteed. As such, some degree of noise or misalignment may persist despite the technical completeness of the preprocessing step. 

For the \textit{POS tagging} task, a critical challenge emerges from the translation of datasets that were originally annotated with Universal POS (UPOS) tags at the word level. The translation model used, \textit{Helsinki-NLP/Opus-MT}, is optimised for sentence-level translation. Although this improves fluency and reduces hallucinations, it also means that translated sentences no longer have a one-to-one correspondence with the original tokenised forms. As a result, the original UPOS annotations become misaligned or unusable. To overcome this, we opt to recompute POS tags directly on the translated English text using the \textit{spaCy} library. Each translated sentence is processed with spaCy's pretrained English pipeline, which outputs POS tags for each token based on linguistic analysis. This ensures that the number of tokens matches the number of predicted POS tags, restoring alignment, and making the dataset suitable for supervised training.

\subsection{Translation Models}
To translate the non-English datasets into English, we use models from the OPUS-MT project \citep{ct126}. OPUS-MT provides open-source neural machine translation models trained on large-scale parallel corpora from the OPUS bitext repository. For the purposes of this study, OPUS-MT offers an important practical advantage: dedicated models are available for all language pairs considered here (Bulgarian--English, Chinese--English, Dutch--English, Italian--English, and Russian--English), enabling a consistent translation pipeline across all experiments. 

The choice of OPUS-MT was motivated by the central aim of this study: to evaluate a resource-light, transparent, and reproducible translate-train strategy rather than to benchmark state-of-the-art machine translation systems. Commercial systems such as DeepL \citep{deepl} and Google Translate often achieve strong performance in terms of fluency, contextual adequacy, and idiomatic expression, but they also introduce practical constraints related to cost, data sharing, API dependence, version instability, and reproducibility. Similarly, recent multilingual or instruction-tuned large language models provide increasingly strong translation capabilities, but their performance can depend on prompting strategies, decoding parameters, model access, and computational resources. 

OPUS-MT was therefore selected as a practical and reproducible translation component. It can be executed locally within the Hugging Face ecosystem, avoids reliance on commercial APIs, and allows the full preprocessing and fine-tuning pipeline to be replicated across computing environments. This is particularly relevant for low-resource and resource-constrained settings, where access to large proprietary models or paid translation systems may be limited. 

Previous work supports the use of OPUS-MT for applied text-analysis pipelines. In a large-scale comparison of OPUS-MT, Google Translate, and DeepL, Licht et al. found no statistically significant downstream differences for topic modelling and text classification, with average performance differences below 0.01 F1 points and highly similar topic correlations with human-translated benchmarks across 10 European source languages \cite{econtribute2024}. Although these findings do not necessarily generalise to all tasks considered in the present study, especially token-level tasks such as POS tagging and named entity recognition, they indicate that OPUS-MT can provide a competitive and reproducible basis for downstream NLP experiments. 

To further contextualise this choice, we conducted a small qualitative inspection of Dutch sentiment-analysis examples translated with several open multilingual alternatives. This inspection suggested that some larger multilingual models produced more unstable translations in this setting, including untranslated fragments, semantic shifts, omissions, and overly literal renderings. We do not treat this inspection as a full machine translation benchmark, but as additional support for using OPUS-MT as the translation component in a resource-light and reproducible experimental pipeline. A systematic comparison of translation systems, including commercial systems and task-specific LLM-based translation models, remains an important direction for future work. 

\section{Experiments}
We evaluate the effectiveness of translation-based fine-tuning by comparing language-specific BERT models on original texts with an English BERT model fine-tuned on their translated counterparts. Our experiments cover Bulgarian, Chinese, Dutch, Italian, and Russian across six NLP tasks. As described in Section~2.2, we use OPUS-MT to translate all non-English datasets into English. Fine-tuning and hyperparameter tuning are performed using the BERT-base model \citep{bert}, which offers a good balance between efficiency and accuracy, requiring fewer computational resources than BERT-Large while still achieving strong results across many benchmarks \citep{ct75}. Depending on the task, we employ either the \textit{cased} or \textit{uncased} variant. The following subsection details the language-specific BERT models used as baselines for the original texts.

\subsection{Non-English BERT Models}
To establish native-language baselines, we run experiments on the original, non-translated datasets using language-specific BERT-family models. All selected models are publicly available through Hugging Face and were chosen to provide comparable encoder-based baselines across languages. Where possible, we selected BERT-base checkpoints in both cased and uncased variants.

For Bulgarian, we use \textit{bert-web-bg} and the uncased \textit{bert\_bg\_lit\_web\_base\_uncased} model by AIaLT-IICT, both of which are publicly available Bulgarian BERT-based checkpoints suitable for downstream fine-tuning. For Dutch, we rely on BERTje \citep{bertje}, a well-established Dutch BERT-base model that was shown to outperform multilingual BERT on several Dutch downstream tasks. Because BERTje does not provide a directly corresponding uncased Dutch BERT-base checkpoint, we use the same model in both the cased and uncased experimental settings. This choice preserves architectural comparability across experiments, although we acknowledge that newer Dutch encoder models, such as RobBERT, may offer stronger performance but follow a different RoBERTa-based architecture. For Italian, we used the cased and uncased models released by the \textit{dbmdz} team, trained on large Italian corpora and shown to outperform mBERT and XLM-R. For Chinese, we employ Google’s official \textit{bert-base-chinese}, which serves as both a cased and uncased model since the language does not mark case. For Russian, we use \textit{ruBERT} (DeepPavlov) for the cased setting and \textit{deepvk/bert-base-uncased} for the uncased setting, both widely adopted in Russian NLP. A summary of all models and their Hugging Face identifiers is provided in Table~\ref{tab:lang_bert}.
\begin{table*}[t]
\centering
\caption{Hugging Face identifiers of the BERT models used in the language-specific experiments.}
\label{tab:lang_bert}

\small
\setlength{\tabcolsep}{4pt}
\renewcommand{\arraystretch}{1.15}

\begin{tabularx}{\textwidth}{@{}lXX@{}}
\toprule
\textit{Language} & \textit{Cased} & \textit{Uncased} \\
\midrule
Dutch &
\texttt{GroNLP/bert-base-dutch-cased} &
\texttt{GroNLP/bert-base-dutch-cased} \\
Italian &
\texttt{dbmdz/bert-base-italian-cased} &
\texttt{dbmdz/bert-base-italian-uncased} \\
Bulgarian &
\texttt{usmiva/bert-web-bg} &
\texttt{AIaLT-IICT/bert\_bg\_lit\_web\_base\_uncased} \\
Russian &
\texttt{DeepPavlov/rubert-base-cased} &
\texttt{deepvk/bert-base-uncased} \\
Chinese &
\texttt{google-bert/bert-base-chinese} &
\texttt{google-bert/bert-base-chinese} \\
\bottomrule
\end{tabularx}
\end{table*}

\subsection{Hyperparameter Tuning}
We use Optuna \citep{ct3} for automated hyperparameter optimisation based on Bayesian search. The tuned parameters include learning rate, batch size, number of training epochs, and weight decay. Hyperparameter tuning is applied to the QA, HSD, POS tagging, and NER tasks, selecting the configuration with the highest validation F1-score. For NLI and SA, we use fixed hyperparameters: learning rate $5\times10^{-5}$, batch size 16, 3 epochs, and weight decay 0.01, due to their larger dataset sizes and infrastructural constraints.

\subsection{Fine-tuning}
We fine-tune all BERT models using Hugging Face's \textit{Trainer} API, which simplifies the process of training and evaluating transformer-based models. With the \textit{Trainer} class, we can automate forward passes, backpropagation, optimisation, evaluation, and checkpoint management. This training configuration allows us to maintain a clear and modular training pipeline while ensuring reproducibility and consistency between experiments. All models employ Hugging Face tokenisers and custom PyTorch Dataset classes for input preparation. Training, validation and test sets are created using an 80/10/10 split when a standard split is not provided by the dataset creators. We use the same random seed to ensure reproducibility.

\subsection{Evaluation}
To ensure a fair and comprehensive comparison between the translation-based and native-language models, we adopt appropriate evaluation metrics according to the task. For classification tasks, we report standard metrics including accuracy, precision, recall, and F1-score, providing a balanced view of performance across majority and minority classes. In highly imbalanced datasets, such as those in hate speech detection and sentiment analysis, the F1-score is emphasised, as it better reflects performance on under-represented categories. 

Although all metrics are reported for completeness, the F1-score is used as the principal basis of comparison across tasks and languages. This choice reflects its ability to balance precision and recall, making it particularly informative in imbalanced classification settings and sequence-labelling tasks where errors are not evenly distributed. Although the F1-score was chosen as the principal metric for comparability across tasks, we note that alternative F$_\beta$-scores (e.g., F0.25) could be considered in contexts where precision is more critical than recall.

For token-level sequence labelling tasks (POS tagging and NER), the evaluation is based on token-level precision, recall, and F1-score, following standard CoNLL evaluation practices \citep{conll}. These metrics capture the models' abilities to assign correct labels at the subword level, with special handling to ensure alignment between the original labels and BERT’s WordPiece tokenization.

For span-based tasks such as QA, we employ Exact Match (EM) and F1-score. EM measures the proportion of predictions that match the gold-standard answer span exactly, while F1 accounts for the partial overlap between the predicted and reference answers, thus providing a more nuanced reflection of the quality of the answer.

Finally, to summarise relative performance across all tasks and languages, we compute the difference in F1-scores between translation-based and native-language models. The results are categorised as better, worse or comparable, and the results are considered comparable if the performance gap lies below 3 percentage points on either side. With these evaluation criteria established, we now turn to the experimental results, analysing performance differences between translation-based and native-language BERT models across tasks and languages.

\section{Results}
In this section, we evaluate to which extent translation can serve as a viable alternative to language-specific modelling. The results are presented by task, following an increasing order of linguistic and semantic complexity. We begin with Part-of-Speech (POS) tagging and Named Entity Recognition (NER), which primarily test syntactic and token-level accuracy. We then move to Sentiment Analysis (SA) and Hate Speech Detection (HSD), which involve more abstract, lexical, and affective interpretation. Finally, we consider Natural Language Inference (NLI) and Question Answering (QA), which require higher-level reasoning and contextual understanding. Within each task, languages are ordered by their linguistic similarity to English. This ordering facilitates interpretation of cross-linguistic trends, as translation-based models are expected to perform more closely to native models for languages that are typologically similar to English, while larger gaps may emerge for more distant languages. 

Throughout this section, we treat the native-language BERT model as the reference baseline for each language–task pair. The translation-based model, created by translating the input into English and fine-tuning English BERT, is evaluated against this baseline. Therefore, terms such as ‘performance drop,’ ‘improvement,’ or ‘comparable results’ refer to changes in performance relative to this baseline.

\subsection{Part-of-Speech Tagging}
Table~\ref{tab:pos_res} presents the results for the POS tagging task. It is important to interpret the POS tagging results differently from the other tasks. Because POS labels for the translated datasets were automatically recomputed on the English translations, these resuslts should be interpreted as evaluating the quality and usability of the translated silver-labelled POS pipeline rather than as a direct comparison to the original gold-labelled POS task. Under this setup, the translated Dutch and Chinese datasets yielded higher F1-scores than their original-language counterparts, while Italian and Russian showed clear decreases and Bulgarian remained largely stable. These patterns suggest that translated English POS data can support effective model training in some cases, but the results are not directly comparable to tasks where the original gold labels are preserved. 
\begin{table}[h] \centering \scriptsize \setlength{\tabcolsep}{3pt} \renewcommand{\arraystretch}{1.08} \caption{Results on the test sets of the POS tagging datasets obtained after fine-tuning. The $\triangle$F1 column represents the performance change when using translated data.} \label{tab:pos_res} \begin{tabular}{llccccc} \toprule \textit{Language} & \textit{Data} & \textit{Acc.} & \textit{Prec.} & \textit{Rec.} & \textit{F1} & \textit{$\triangle$F1} \\ \midrule Dutch & Translated & 0.9643 & 0.8505 & 0.9138 & 0.8699 & +8.68 \\ & Original & 0.8442 & 0.7741 & 0.8116 & 0.7831 & \\ \addlinespace[0.2em] Italian & Translated & 0.7492 & 0.8057 & 0.8258 & 0.7881 & -14.47 \\ & Original & 0.9621 & 0.9256 & 0.9411 & 0.9328 & \\ \addlinespace[0.2em] Bulgarian & Translated & 0.9745 & 0.9114 & 0.9528 & 0.9285 & -1.78 \\ & Original & 0.9764 & 0.9144 & 0.9806 & 0.9463 & \\ \addlinespace[0.2em] Russian & Translated & 0.8681 & 0.8052 & 0.8445 & 0.8122 & -8.13 \\ & Original & 0.9407 & 0.8678 & 0.9338 & 0.8935 & \\ \addlinespace[0.2em] Chinese & Translated & 0.9819 & 0.9230 & 0.9573 & 0.9369 & +13.31 \\ & Original & 0.8569 & 0.7637 & 0.8791 & 0.8038 & \\ \bottomrule \end{tabular} \end{table}

\subsection{Named Entity Recognition}
Table \ref{tab:ner_res} presents the results for the NER task. Overall, the models achieve strong performance on the original datasets for Bulgarian, Italian, and Russian. For Bulgarian and Italian, the models trained on translated data show some performance degradation relative to the original-language models, but the resulting F1-scores remain reasonably high. This indicates that the models maintain a substantial degree of effectiveness on the translated datasets, though with reduced reliability. In contrast, the performance on the translated Russian data declines sharply, suggesting that translation introduces substantial noise that negatively affects the NER task for this language.

For Dutch, the models achieve moderate performance in both settings, with the translated-data model showing a slight improvement, indicating relative robustness to translation-induced variation. For Chinese, the performance decreases notably after translation, indicating challenges in modelling NER after translating Chinese to English.

\begin{table}[h] \centering \scriptsize \setlength{\tabcolsep}{3pt} \renewcommand{\arraystretch}{1.08} \caption{Results on the test sets of the NER datasets obtained after fine-tuning. The $\triangle$F1 column represents the performance change when using translated data.} \label{tab:ner_res} \begin{tabular}{llccccc} \toprule \textit{Language} & \textit{Data} & \textit{Acc.} & \textit{Prec.} & \textit{Rec.} & \textit{F1} & \textit{$\triangle$F1} \\ \midrule Dutch & Translated & 0.9363 & 0.7439 & 0.7982 & 0.7701 & +2.75 \\ & Original & 0.8961 & 0.7011 & 0.7892 & 0.7426 & \\ \addlinespace[0.2em] Italian & Translated & 0.9141 & 0.7421 & 0.8030 & 0.7713 & -10.29 \\ & Original & 0.9486 & 0.8526 & 0.8969 & 0.8742 & \\ \addlinespace[0.2em] Bulgarian & Translated & 0.9291 & 0.6906 & 0.8388 & 0.7575 & -14.50 \\ & Original & 0.9513 & 0.8829 & 0.9230 & 0.9025 & \\ \addlinespace[0.2em] Russian & Translated & 0.8172 & 0.2685 & 0.6446 & 0.3791 & -50.05 \\ & Original & 0.9450 & 0.8593 & 0.9008 & 0.8796 & \\ \addlinespace[0.2em] Chinese & Translated & 0.8296 & 0.4063 & 0.7317 & 0.5225 & -7.54 \\ & Original & 0.8578 & 0.4803 & 0.7917 & 0.5979 & \\ \bottomrule \end{tabular} \end{table}

\subsection{Sentiment Analysis}
Table \ref{tab:sa_res} presents the results for the SA task. Performance varies heavily across languages. For Bulgarian, the model trained on Bulgarian text shows moderate performance, with an accuracy of 79.5\% and an F1-score of 66.6\%. The translated approach shows a decline across all metrics, particularly in precision, indicating a higher rate of false positives.

In Chinese, the respective model achieves near-perfect accuracy and F1-score in the original-language setting. The translated version shows a notable decrease, but performance remains high overall. 

Dutch also yields very high performance, with the models trained on original and translated input achieving nearly identical results. The dataset is binary and balanced, which supports consistent performance across both configurations.

For the Italian dataset, results are more nuanced. The model on translated input achieves slightly higher accuracy, but both models converge at a similar F1-score, indicating that despite high overall accuracy, the models may have difficulty with minority sentiment classes. This reflects a common effect of class imbalance, where accuracy overestimates true performance.

Finally, for Russian, the models show solid performance in both configurations, and the translated version exhibits only a modest decrease across all metrics, indicating that the model trained on translated data is effective for SA in this context.

\begin{table}[h] \centering \scriptsize \setlength{\tabcolsep}{3pt} \renewcommand{\arraystretch}{1.08} \caption{Results on the test sets of the SA datasets obtained after fine-tuning. The $\triangle$F1 column represents the performance change when using translated data.} \label{tab:sa_res} \begin{tabular}{llccccc} \toprule \textit{Language} & \textit{Data} & \textit{Acc.} & \textit{Prec.} & \textit{Rec.} & \textit{F1} & \textit{$\triangle$F1} \\ \midrule Dutch & Translated & 0.9283 & 0.9284 & 0.9282 & 0.9283 & -0.07 \\ & Original & 0.9293 & 0.9295 & 0.9290 & 0.9290 & \\ \addlinespace[0.2em] Italian & Translated & 0.8341 & 0.5745 & 0.5861 & 0.5603 & -0.05 \\ & Original & 0.7993 & 0.5063 & 0.6815 & 0.5608 & \\ \addlinespace[0.2em] Bulgarian & Translated & 0.7620 & 0.6332 & 0.6432 & 0.6380 & -2.76 \\ & Original & 0.7945 & 0.6963 & 0.6632 & 0.6656 & \\ \addlinespace[0.2em] Russian & Translated & 0.7417 & 0.7621 & 0.7414 & 0.7439 & -3.10 \\ & Original & 0.7719 & 0.7833 & 0.7715 & 0.7749 & \\ \addlinespace[0.2em] Chinese & Translated & 0.8294 & 0.8297 & 0.8293 & 0.8293 & -15.49 \\ & Original & 0.9842 & 0.9846 & 0.9843 & 0.9842 & \\ \bottomrule \end{tabular} \end{table}

\subsection{Hate Speech Detection}
Table \ref{tab:hsd_res} presents the results for the HSD task in the five languages considered. Overall, these results reveal varying levels of success in using BERT to detect hate speech, with performance largely dependent on the characteristics of the language and dataset. For Bulgarian, both models achieve very high accuracy (above 97.8\%), indicating strong overall classification capability. However, the considerably lower precision, recall, and F1-scores (all below 0.74) indicate difficulty in correctly identifying hate speech instances, likely in part due to class imbalance and the limited ability to predict the minority class.

The largest performance gap is observed for Chinese, where the model on translated input substantially underperforms the original-language BERT by more than 7\% across all metrics. The translated model reaches an F1-score of 0.8303, compared to 0.9063 in the model trained on Chinese texts. For Dutch, the original-language model demonstrates the best performance across all metrics, achieving an F1-score of 0.9829, the highest among all evaluated hate speech datasets. The translated model also performs well, indicating that both configurations are effective. The strong results, especially in recall on the hateful class, show that both models can reliably identify hate speech in this dataset, regardless of translation.

In the case of Italian, both models show more modest results, with F1-scores below 0.79. Although the original-language model performs slightly better (by about 3–4\% in recall and F1), the overall performance of the translated model indicates moderate effectiveness. The models appear to capture hate speech patterns to some extent, but with noticeable limitations in distinguishing subtle cases.

For Russian, the differences between the translated and original versions are relatively small, with the translated model reporting a decline of only about 1.5\% across most metrics. Both versions yield F1-scores in the range of 0.86–0.88, indicating solid performance.
\begin{table}[h] \centering \scriptsize \setlength{\tabcolsep}{3pt} \renewcommand{\arraystretch}{1.08} \caption{Results on the test sets of the HSD datasets obtained after fine-tuning. The $\triangle$F1 column represents the performance change when using translated data.} \label{tab:hsd_res} \begin{tabular}{llccccc} \toprule \textit{Language} & \textit{Data} & \textit{Acc.} & \textit{Prec.} & \textit{Rec.} & \textit{F1} & \textit{$\triangle$F1} \\ \midrule Dutch & Translated & 0.9469 & 0.9574 & 0.9081 & 0.9291 & -5.38 \\ & Original & 0.9867 & 0.9911 & 0.9755 & 0.9829 & \\ \addlinespace[0.2em] Italian & Translated & 0.7596 & 0.7522 & 0.7511 & 0.7516 & -3.71 \\ & Original & 0.8002 & 0.8008 & 0.7833 & 0.7887 & \\ \addlinespace[0.2em] Bulgarian & Translated & 0.9817 & 0.7163 & 0.6889 & 0.7016 & -2.36 \\ & Original & 0.9782 & 0.7104 & 0.7424 & 0.7252 & \\ \addlinespace[0.2em] Russian & Translated & 0.8797 & 0.8560 & 0.8685 & 0.8617 & -1.34 \\ & Original & 0.8911 & 0.8683 & 0.8835 & 0.8751 & \\ \addlinespace[0.2em] Chinese & Translated & 0.8303 & 0.8303 & 0.8307 & 0.8303 & -7.60 \\ & Original & 0.9064 & 0.9065 & 0.9066 & 0.9063 & \\ \bottomrule \end{tabular} \end{table}

\subsection{Natural Language Inference}
Table \ref{tab:nli_res} shows the results for the NLI task, which demonstrate relatively stable model performance across both original and translated datasets. For Bulgarian, the models perform well on both original and translated input. The difference is minimal, indicating the models can maintain robust inference capabilities regardless of whether the data is in the original language or translated.
In Chinese, the performance is slightly lower overall but remains reasonably strong, suggesting that the models can still effectively perform NLI, although with some sensitivity to translation quality or linguistic complexity.
Dutch also shows strong and consistent results, and for Italian, the model trained on the translated dataset actually outperforms the model trained on the original texts. Although both Italian and Dutch yield lower scores than other languages, small translation improvement implies that translation may help mitigate some limitations in the original data quality or structure. 
Russian is another case where the translated version marginally outperforms the original, again reflecting resilience to translating in the NLI setting and suggesting effective generalisation.

\begin{table}[h] \centering \scriptsize \setlength{\tabcolsep}{3pt} \renewcommand{\arraystretch}{1.08} \caption{Results on the test sets of the NLI datasets obtained after fine-tuning. The $\triangle$F1 column represents the performance change when using translated data.} \label{tab:nli_res} \begin{tabular}{llccccc} \toprule \textit{Language} & \textit{Data} & \textit{Acc.} & \textit{Prec.} & \textit{Rec.} & \textit{F1} & \textit{$\triangle$F1} \\ \midrule Dutch & Translated & 0.8333 & 0.8166 & 0.8377 & 0.8248 & -0.42 \\ & Original & 0.8394 & 0.8100 & 0.8558 & 0.8290 & \\ \addlinespace[0.2em] Italian & Translated & 0.6451 & 0.6434 & 0.6440 & 0.6434 & +2.27 \\ & Original & 0.6210 & 0.6256 & 0.6207 & 0.6207 & \\ \addlinespace[0.2em] Bulgarian & Translated & 0.7771 & 0.7776 & 0.7771 & 0.7773 & -0.75 \\ & Original & 0.7847 & 0.7868 & 0.7849 & 0.7848 & \\ \addlinespace[0.2em] Russian & Translated & 0.7702 & 0.7707 & 0.7702 & 0.7704 & +1.13 \\ & Original & 0.7599 & 0.7600 & 0.7600 & 0.7591 & \\ \addlinespace[0.2em] Chinese & Translated & 0.7390 & 0.7384 & 0.7389 & 0.7382 & -3.47 \\ & Original & 0.7723 & 0.7749 & 0.7722 & 0.7729 & \\ \bottomrule \end{tabular} \end{table}

\subsection{Question Answering}
Table \ref{tab:qa_res} presents the results for the QA task, evaluated using EM and F1-score. In general, the results reveal significant variability in performance across languages and datasets, driven by factors such as dataset size, context complexity, and preprocessing challenges related to answer alignment. Here, context complexity refers to the structural and semantic difficulty of the passages provided for question answering. Longer, information-dense, or syntactically complex contexts increase the challenge of locating answer spans, particularly after translation, where meaning or alignment can be partially distorted.
For Bulgarian, the performance is low across metrics, regardless of whether the original or translated version is used. The highest EM recorded is only 21.1\%, and the F1-score reaches 33.3\%. These modest scores suggest that the models struggle with the task on this dataset, possibly due to the relatively small size of the dataset (3,349 examples) and the fact that the dataset was originally created for multiple-choice answers. These factors likely make it difficult for the models to locate and extract the answer spans accurately. The low EM and F1-scores reflect this challenge, although we do not measure the average positional error directly; further span alignment analysis could clarify whether the errors are due to mislocation or semantic mismatch.
For Chinese, the results are noticeably better. The original-language model achieves an EM of 0.461 and an F1-score of 0.469, demonstrating moderate success. In contrast, the translation-based approach suffers a noticeable drop in EM (to 0.295), although the F1-score remains relatively close (0.425). 
For Dutch, the translated model achieved an EM of 0.214 and an F1-score of 0.509. The original-language model, by contrast, produces no exact matches, highlighting alignment or formatting issues in the original version. The translated model demonstrates reasonable capability for Dutch QA despite the limited size of the dataset.
In Italian, the results are especially strong for the translated version, which achieves the second-highest F1-score overall (0.701) and a high EM of 0.583. The original-language model performs substantially worse. The translated setup, supported by the English QA span inference with distilBERT, enables effective model training and prediction.
The Russian dataset yields the highest performance among all evaluated datasets, with the translated model achieving an EM score of 0.652 and an F1-score of 0.763. The original-language model also attains strong results, indicating that the dataset is well-structured and particularly suitable for span-based QA evaluation.
\begin{table}[h] \centering \scriptsize \setlength{\tabcolsep}{3pt} \renewcommand{\arraystretch}{1.08} \caption{Results on the test sets of the QA datasets obtained after fine-tuning. The $\triangle$F1 column represents the performance change when using translated data.} \label{tab:qa_res} \begin{tabular}{llccc} \toprule \textit{Language} & \textit{Data} & \textit{EM} & \textit{F1} & \textit{$\triangle$F1} \\ \midrule Dutch & Translated & 0.2143 & 0.5095 & +4.50 \\ & Original & 0.0000 & 0.4644 & \\ \addlinespace[0.2em] Italian & Translated & 0.5829 & 0.7012 & +13.34 \\ & Original & 0.0869 & 0.5678 & \\ \addlinespace[0.2em] Bulgarian & Translated & 0.2112 & 0.2992 & -3.38 \\ & Original & 0.1618 & 0.3330 & \\ \addlinespace[0.2em] Russian & Translated & 0.6521 & 0.7634 & +2.14 \\ & Original & 0.5358 & 0.7420 & \\ \addlinespace[0.2em] Chinese & Translated & 0.2948 & 0.4245 & -4.45 \\ & Original & 0.4607 & 0.4690 & \\ \bottomrule \end{tabular} \end{table}

\subsection{Overall Performance Overview}
Table~\ref{tab:ovw_res} summarises the results of the translation-based model relative to native-language baselines across all tasks and languages. Each language-task combination is classified into three categories per task: translation-based results were worse than native-language BERT ($\times$), comparable ($\sim$), or better ($\checkmark$). Results are considered comparable when the translation-based approach achieves a performance within 3 percentage points. For most tasks this refers to F1-score, while for question answering it refers to the span-level F1-score.

For interpretability, the tasks in Table~\ref{tab:ovw_res} are ordered by increasing task complexity; from structurally orientated sequence-labelling tasks (POS tagging, NER) to semantically demanding classification and inference tasks (HSD, NLI, QA). Similarly, the languages are arranged approximately by typological proximity to English, from Dutch (closest) to Chinese (most distant), allowing clearer comparison of cross-linguistic patterns.

Across all 30 language--task combinations, the translation-based approach achieved comparable or better performance in 16 cases (53.3\%). However, the POS tagging results require a more cautious interpretation, because POS labels for the translated datasets were automatically recomputed on the English translations rather than projected from the original gold-standard annotations. The translated POS setting should therefore be understood as a silver-label experiment rather than a direct gold-label comparison. Under this stricter interpretation, the translation-based approach remains comparable or better in 13 out of 25 settings (52.0\%). 

A further summary excluding both token-level tasks, POS tagging and NER, shows that translation-based fine-tuning is comparable or superior in 12 out of 20 sentence-level or span-based settings (60.0\%). This supports the broader pattern observed in the task-specific results: translation-based fine-tuning is most reliable for tasks where labels are attached to complete sequences or answer spans, such as SA, HSD, NLI, and QA, and less reliable for tasks that depend on preserving token boundaries and word-level annotations. Improvements are most frequent in QA and NLI, while performance drops are more common in NER and HSD. Language-wise, Dutch shows the highest rate of comparable or better outcomes, while Chinese shows the fewest such cases. These cross-task trends highlight both the potential and the limitations of translation-based fine-tuning. In the following Discussion section, we examine the linguistic, task-specific, and methodological factors that may explain these patterns. 
\begin{table}[h] \centering \scriptsize \setlength{\tabcolsep}{4pt} \renewcommand{\arraystretch}{1.12} \caption{Performance of translation-based models compared to native-language models across tasks and languages. Ticks ($\checkmark$) indicate better performance, crosses ($\times$) indicate worse performance, and tildes ($\sim$) indicate comparable results, defined as less than 3 percentage points difference in the primary evaluation metric.} \label{tab:ovw_res} \begin{tabular}{lccccc} \toprule \textit{Task} & \textit{Dutch} & \textit{Italian} & \textit{Bulgarian} & \textit{Russian} & \textit{Chinese} \\ \midrule POS & $\checkmark$ & $\times$ & $\sim$ & $\times$ & $\checkmark$ \\ NER & $\sim$ & $\times$ & $\times$ & $\times$ & $\times$ \\ SA & $\sim$ & $\sim$ & $\sim$ & $\times$ & $\times$ \\ HSD & $\times$ & $\times$ & $\sim$ & $\sim$ & $\times$ \\ NLI & $\sim$ & $\sim$ & $\sim$ & $\sim$ & $\times$ \\ QA & $\checkmark$ & $\checkmark$ & $\times$ & $\sim$ & $\times$ \\ \bottomrule \end{tabular} \vspace{0.4em} \begin{minipage}{\linewidth} \scriptsize \textit{Note.} POS = Part-of-Speech Tagging; NER = Named Entity Recognition; SA = Sentiment Analysis; HSD = Hate Speech Detection; NLI = Natural Language Inference; QA = Question Answering. \end{minipage} \end{table}

\section{Discussion}
The results presented in Section~4 provide a comprehensive view of how translation-based fine-tuning compares to native-language BERT models across multiple tasks and languages. Overall, the findings suggest that translating data into English and fine-tuning an English BERT model can serve as a practical and resource-light alternative under specific conditions, rather than as a general replacement for language-specific models, multilingual encoders, or LLM-based approaches. Its success depends on the interaction between translation fidelity, task type, label structure, dataset characteristics, and linguistic distance. This section contextualises these findings within broader NLP research and discusses their implications for cross-lingual transfer, reproducibility, and future research.

\subsection{Cross-linguistic Trends}
The cross-linguistic patterns reveal a clear gradual variation in performance linked to typological proximity to English. Translation-based models achieved the strongest results for Dutch, which shares numerous structural features with English, including relatively fixed word order and shared Germanic roots. These factors likely helped English BERT's pretrained representations to generalise more effectively. This observation is consistent with previous work by Wu and Dredze, who demonstrated that cross-lingual transfer efficiency is significantly influenced by linguistic similarity \cite{ct144}. However, typological proximity may not have directly influenced the models' learning capacity after translation, but rather indirectly through translation quality. Prior evaluations of OPUS-MT \citep{ctTiedemann2021, ctOPUS2022} have shown that translation accuracy is generally higher for typologically related languages and degrades for more distant pairs, suggesting that linguistic similarity improves both translation accuracy and downstream model performance.

Although typological proximity clearly helps, it is not the sole determinant of success. The Slavic languages Bulgarian and Russian still showed competitive performance after translation in several tasks, despite their greater morphological complexity and freer word order. This indicates that translation-based fine-tuning can bridge moderate typological distances when translation quality is sufficient. Interestingly, Russian frequently maintained strong performance across multiple tasks after translation, suggesting that the robustness of the source datasets (e.g., SberQuAD for QA and XNLI-ru for NLI) and the strong generalisation capacity of English BERT can partially mitigate typological distance. By contrast, Bulgarian, which relied on smaller and more imbalanced datasets, showed slightly less stability after translation, reinforcing the role of data quality and sample size in the success of cross-lingual transfer.

At the other end of the proximity spectrum, Chinese consistently underperformed in the translation-based setup. This outcome can be attributed to several factors: (1) substantial structural and typological distance from English, including differences in word segmentation, syntax, and morphology; (2) potential translation noise introduced during preprocessing, especially for token-level tasks; and (3) the presence of a possibly strong monolingual baseline in \textit{bert-base-chinese}, which captures language-specific semantics more effectively. Together, these elements likely made it difficult for English BERT to fully replicate the performance of the native Chinese model.

In general, these findings suggest that typological proximity facilitates, but does not fully determine, the success of translation-based fine-tuning. Our experiments show that the interaction between the linguistic distance, the quality of the data set and the task type could jointly shape the performance. Languages that are moderately distant from English but supported by high-quality datasets can still achieve competitive results, highlighting that translation-based approaches are not strictly limited to languages closely related to English, but depend on a balance between linguistic and data-driven factors.

\subsection{Task-specific Patterns}
The results reveal that the effectiveness of the translation-based approach varies considerably between task types, reflecting differences in how sensitive each task is to linguistic nuance, how its data is annotated, and how strictly label alignment must be preserved after translation. Overall, translation-based fine-tuning was most reliable for tasks where labels are attached to complete sequences or answer spans, such as NLI and QA, and less reliable for tasks that depend on preserving word-level boundaries, entity spans, or culturally specific lexical cues. 

For NLI and QA, the translation-based approach often produced comparable or better results than the native-language baselines. A likely explanation is that these tasks depend primarily on sentence- or passage-level semantic relations, such as the relation between a premise and hypothesis or between a question and its supporting context. These relations can often remain intact after translation, even when individual word choices or sentence structures change. However, QA also introduces an additional methodological challenge: answer spans must be re-identified in the translated context. As a result, QA performance should be interpreted not only as a measure of model performance, but also as reflecting the reliability of the answer-span realignment procedure. 

In contrast, tasks such as SA and HSD were more vulnerable to translation noise and domain mismatch. These tasks rely on subtle lexical and cultural cues; such as figurative expressions, irony, or slang that carry emotional or social meaning. During translation, these nuances are frequently softened, neutralised, or interpreted differently, which can alter the original intent or tone of the text. In some cases, meaning may even be reversed: ironic or sarcastic expressions that convey negativity in the original language can be translated literally, causing the models to possibly misinterpret them as positive statements. As a result, translated data may not convey the emotional intensity or offensiveness that models need to detect, reducing their effectiveness in these settings.

The token-level tasks require the most cautious interpretation. NER performance was generally weaker after translation, which is consistent with the fact that translation can alter entity surface forms, token boundaries, word order, and entity spans. These changes make it difficult to preserve a direct correspondence between the translated text and the original gold-standard annotations. POS tagging differs further from the other tasks because the original POS labels could not be directly preserved after sentence-level translation. Instead, POS labels were recomputed on the translated English text using an automatic English POS tagger. For this reason, the POS findings should be interpreted as evidence about the feasibility of constructing a usable translated POS pipeline, not as evidence that translation preserves original POS annotations without loss. 

Taken together, these task-specific patterns suggest that translation-based fine-tuning is most suitable for tasks in which the relevant information is expressed at the sentence, passage, or answer-span level, and less suitable for tasks where the labels are tightly coupled to language-specific surface forms, token boundaries, or culturally situated expressions. This distinction helps explain why the approach remains promising for resource-light cross-lingual transfer, while also highlighting the methodological limits of using translation as a general-purpose bridge across all NLP tasks. 

\subsection{Dataset Quality and Label Distribution}
The results of our experiments suggest that dataset characteristics, while relevant, played a secondary role compared to typological and linguistic factors in determining translation-based performance. This conclusion is supported by several cross-linguistic patterns in our results: languages with relatively small or imbalanced datasets, such as Dutch, often achieved strong performance, while Chinese, which had large and well-balanced corpora, consistently underperformed. These discrepancies indicate that dataset size and label distribution alone do not account for the observed performance differences, suggesting that linguistic proximity and translation fidelity exert a stronger and more systematic influence on model outcomes. 

As shown in Table~\ref{tab:datasets_overview} and Appendix A, dataset size and label balance varied widely across tasks and languages, introducing differences in task difficulty and robustness. For instance, while the Italian QA dataset (QA-ITA-200k) and Russian NLI corpus (XNLI-ru) contained hundreds of thousands of examples, Dutch QA and HSD datasets were limited to only a few hundred and a few thousand samples, respectively. Similarly, several datasets, particularly Bulgarian HSD and SA, showed pronounced class imbalance, with minority classes representing less than 10\% of the data.

Despite these disparities, dataset size and label distribution did not consistently predict model performance. Dutch, for example, frequently achieved strong results despite comparatively small datasets, suggesting that typological proximity and translation quality had a stronger impact than dataset size. Furthermore, Chinese tasks, which were supported by large and balanced datasets, consistently underperformed, further indicating that linguistic distance is more important in this context than data availability. For Russian, the high quality and well-structured datasets themselves likely contributed to the strong results across tasks, mitigating, but not eliminating, the challenges posed by the language’s morphological complexity and typological distance from English.

Overall, these results suggest that dataset quality and label balance shape performance to a degree, but they are not the main determinants of success. Instead, their influence interacts with typological distance and translation fidelity: large, well-balanced datasets can compensate for moderate linguistic differences (as seen in Russian), but they cannot overcome substantial typological or structural divergence (as observed in Chinese). This reinforces that cross-lingual transfer effectiveness depends on the combined effects of data quality, linguistic similarity, and the stability of the translation process.

\subsection{Resource Efficiency}
Beyond performance, an important consideration is computational efficiency. Although this study did not directly evaluate computational resource use, the methodological design itself implies substantial potential savings. Fine-tuning a pre-trained English BERT model is less resource-intensive than pretraining language-specific models from scratch, which typically require hundreds of GPU hours and massive corpora \citep{cit_costnlp, cit_energy, cit_extremebert}. By leveraging existing English models, translation-based fine-tuning avoids this pretraining overhead while enabling cross-lingual generalisation. However, this efficiency gain is partly lowered by the additional inference stage introduced by the translation step. Although this cost is modest relative to pretraining, it can accumulate in large-scale or continuously deployed systems where inference occurs frequently \citep{cit_energyfine, cit_energyfine2}. Therefore, while translation-based fine-tuning offers a resource-efficient alternative to multilingual model training in concept, the exact magnitude of efficiency gains depends on how translation is integrated into practical workflows and the scale at which the system operates. Nevertheless, the approach offers a scalable and useful blueprint for extending NLP technologies to languages with limited digital resources, reducing the overall entry barrier for cross-lingual research and application.
 
\subsection{Limitations and Future Directions}
Several limitations of this study also point to directions for future research. First, the current study focusses on five languages that span a limited subset of the world’s linguistic diversity. Although this selection allowed coverage across major language families and typological distances, expanding the evaluation to include additional and especially under-represented languages would allow for a more comprehensive understanding of how structural and genealogical factors mediate translation-based transfer. This broader scope could also reveal whether the patterns observed here generalise to languages with non-Indo-European origins and/or limited digital presence.

Second, although OPUS-MT was chosen for its open availability and reproducibility, translation quality likely remains an important source of variation. Prior evaluations of OPUS-MT show that performance is strongly influenced by typological distance, with higher translation quality for languages structurally similar to English and lower performance for morphologically rich or distant languages \citep{cit_opusdashboard, cit_loresmt}. However, this study did not explicitly quantify translation quality or control for its impact on model performance. Future work should incorporate translation quality metrics (e.g., BLEU, COMET) or human evaluation to better isolate how much of the performance difference arises from linguistic distance versus translation artefacts. Comparing OPUS-MT with stronger commercial or neural translation systems could also clarify how far translation quality limits the potential of this approach.

A related limitation is that the present study does not constitute a benchmark of machine translation systems. OPUS-MT was selected as a consistent, open-source, and resource-light translation component available for all language pairs considered in this study. This choice aligns with our focus on reproducible translate-train pipelines, but it also means that the results should not be interpreted as evidence that OPUS-MT is the optimal translation system for each language or task. 

Recent work directly comparing OPUS-MT with LLM-based translation systems suggests a nuanced picture: LLMs and instruction-tuned translation models may offer stronger robustness or lexical disambiguation in some settings, but specialised neural MT systems such as OPUS-MT can remain relevant as resource-light translation components in controlled translate-train pipelines, particularly when local execution, low computational cost, transparency, and reproducibility are central requirements \citep{peters2025robustmt,martelli2025dibimt,boyd2025mtera}.

This work compared translation-based fine-tuning and native-language BERT models under a consistent experimental framework. Future studies could build on this by exploring hybrid architectures that integrate translation-based and multilingual approaches. For instance, translation could serve as a preprocessing step before multilingual fine-tuning, potentially combining the strengths of both methods. Similarly, multi-task learning setups that jointly train translation and downstream NLP tasks could reduce translation noise and improve cross-lingual generalisation.

Expanding the scope of evaluation to include more diverse NLP domains could provide valuable insights. Tasks such as summarisation, reading comprehension, and dialogue modelling may respond differently to translation-based transfer, especially when discourse coherence or long-range dependencies play a key role. Moreover, evaluating domain-specific applications such as biomedical, legal, or social media contexts would help assess how translation-based strategies perform in noisy or specialised settings, such as in previous work by Muizelaar et al., which examined translation in a biomedical context \cite{muiz}.

Finally, while this study focused on encoder-based architectures such as BERT, emerging large language models (LLMs) present an alternative avenue. These models excel in zero-shot and few-shot transfer \citep{cit_zero}, but challenges remain around reproducibility \citep{cit_repro, cit_repro2, cit_llmchal, cit_llmchal2}, structured output formats (e.g., token-level predictions) \citep{cit_struct, cit_struct2, cit_llmchal2}, and open access \citep{cit_llmchal, cit_llmchal2}. Exploring hybrid strategies that combine translation with multilingual or LLM-based approaches may provide promising new directions.

More broadly, this study contributes to a growing effort in NLP to develop sustainable and inclusive modelling strategies. By showing that translation-based fine-tuning can achieve competitive results across over half of all tested language-task combinations, it provides evidence that resource-efficient translation is a viable path forward. This directly challenges the assumption that every language requires its own pretrained model, and instead suggests that improving translation systems and fine-tuning pipelines could offer a scalable route to multilingual NLP. 

\section{Conclusions}
This study systematically evaluated the viability of translation-based fine-tuning as an alternative to native-language BERT models across five languages and six NLP tasks. Overall, the results show that translation-based models achieved comparable or better performance in approximately half of all evaluated settings. The strongest gains were observed in QA and NLI, while more pronounced drops occurred in NER and HSD. POS tagging showed promise, but results are methodologically different due to label reconstruction. 

Language-wise, Dutch consistently achieved the highest relative performance, likely reflecting its close typological similarity to English, which may facilitate more reliable translation and smoother model transfer. Russian and Bulgarian produced mixed but competitive outcomes, suggesting that translation-based fine-tuning can bridge moderate typological distances when supported by strong datasets. In contrast, Chinese underperformed across nearly all tasks, reflecting the challenges associated with greater linguistic distance, structural divergence, and potential translation inconsistencies. Importantly, dataset size and label distribution appeared to play a secondary role compared to linguistic and typological factors, as smaller, imbalanced datasets still yielded acceptable results in Dutch.

In summary, translation-based fine-tuning with English BERT emerges as a promising but context-dependent strategy. It performs best for languages that are structurally similar to English and for tasks that rely more on syntactic or relational structure than on fine-grained lexical or culturally specific cues. While it cannot universally replace native-language BERT models, this approach offers a scalable, resource-efficient path for extending NLP to under-represented languages, particularly in contexts where high-quality native models or annotated data are unavailable.

\section*{Declarations}
\begin{itemize}
\item Funding \\
This research did not receive any specific grant from funding agencies in the public, commercial, or not-for-profit sectors.
\item Competing interests\\
None declared.
\item Data and Code availability \\
Our code is available on GitHub, via \url{https://github.com/GiuliaRivets01/Master-Thesis/tree/main}.
\item CRediT authorship contribution statement\\
H.M., M.S., M.H. were responsible for conceptualisation, G.R. performed data curation, the methodology was developed by H.M., G.R., and M.H.. H.M., M.S., and M.H. were responsible for project administration and supervision, H.M, M.S. and M.H. provided resources, G.R. created the software, performed the experiments, validated the results, and created visualisations. The original draft of this manuscript was written by H.M.. Reviewing and editing was done by H.M., G.R., M.H. and M.S.. All authors contributed to the production and proofing of the manuscript. 
\item Generative AI declaration\\
During the preparation of this work, the authors used ChatGPT to support language refinement. The authors reviewed and edited the content as needed and take full responsibility for the content of the submitted manuscript.
\end{itemize}
\typeout{}
\bibliography{cas-refs}

\newpage

\appendix
\setcounter{table}{0}
\renewcommand{\thetable}{A.\arabic{table}}
\section*{Appendices}
\subsection*{A. Label Distributions by Classification Task}
\begin{enumerate}
    \item \textit{Sentiment Analysis}
        \begin{table}[h] \centering \scriptsize \setlength{\tabcolsep}{3pt} \renewcommand{\arraystretch}{1.08} \caption{Dataset and label distribution per language for the SA task.} \label{tab:sa_distribution} \begin{tabular}{lcccc} \toprule \textit{Language} & \textit{Size} & \textit{Positive} & \textit{Neutral} & \textit{Negative} \\ \midrule Dutch & 22,352 & 11,176 (50.0\%) & -- & 11,176 (50.0\%) \\ Italian & 165,294 & 23,556 (14.3\%) & 135,195 (81.8\%) & 6,543 (4.0\%) \\ Bulgarian & 10,198 & 7,638 (74.9\%) & 1,675 (16.4\%) & 885 (8.7\%) \\ Russian & 60,002 & 30,001 (50.0\%) & -- & 30,001 (50.0\%) \\ Chinese & 119,992 & 59,993 (50.0\%) & -- & 59,995 (50.0\%) \\ \bottomrule \end{tabular} \end{table}
    \item \textit{Hate Speech Detection}
    \begin{table}[h] \centering \scriptsize \setlength{\tabcolsep}{4pt} \renewcommand{\arraystretch}{1.08} \caption{Dataset and label distribution per language for the HSD task.} \label{tab:hsd_distribution} \begin{tabular}{lccc} \toprule \textit{Language} & \textit{Size} & \textit{Hateful} & \textit{Non-Hateful} \\ \midrule Dutch & 3,765 & 2,640 (70.1\%) & 1,125 (29.9\%) \\ Italian & 8,602 & 3,569 (41.5\%) & 5,033 (58.5\%) \\ Bulgarian & 102,750 & 1,854 (1.8\%) & 100,896 (98.2\%) \\ Russian & 15,875 & 5,212 (32.8\%) & 10,663 (67.2\%) \\ Chinese & 37,480 & 18,041 (48.1\%) & 19,439 (51.9\%) \\ \bottomrule \end{tabular} \end{table}
    \item \textit{Natural Language Inference}
    \begin{table}[h] \centering \scriptsize \setlength{\tabcolsep}{3pt} \renewcommand{\arraystretch}{1.08} \caption{Dataset and label distribution per language for the NLI task.} \label{tab:nli_distribution} \begin{tabular}{lcccc} \toprule \textit{Language} & \textit{Size} & \textit{Entailment} & \textit{Neutral} & \textit{Contradiction} \\ \midrule Dutch & 9,840 & 2,821 (28.7\%) & 5,595 (56.9\%) & 1,424 (14.5\%) \\ Italian & 34,878 & 11,726 (33.6\%) & 11,591 (33.2\%) & 11,561 (33.1\%) \\ Bulgarian & 400,202 & 133,399 (33.3\%) & 133,400 (33.3\%) & 133,403 (33.3\%) \\ Russian & 408,224 & 136,090 (33.3\%) & 136,051 (33.3\%) & 136,083 (33.3\%) \\ Chinese & 370,188 & 123,309 (33.3\%) & 123,494 (33.3\%) & 123,385 (33.3\%) \\ \bottomrule \end{tabular} \end{table}
\end{enumerate}

\end{document}